\documentclass{article}

\usepackage{arxiv}
\usepackage[numbers]{natbib}

\usepackage[utf8]{inputenc} %
\usepackage[T1]{fontenc}    %
\usepackage{hyperref}       %
\usepackage{url}            %
\usepackage{booktabs}       %
\usepackage{amsfonts}       %
\usepackage{nicefrac}       %
\usepackage{microtype}      %
\usepackage{lipsum}		%
\usepackage{graphicx}
\usepackage{natbib}
\usepackage{doi}
\usepackage{caption}
\usepackage{subcaption}
\usepackage{xspace}

\newcommand\certain{consistent\xspace}
\newcommand\mldc{MLDC\xspace}
\usepackage{makecell}

\newcommand{\mes}[1]{{[}#1{]}}
\newcommand{\s}[2]{#1 $\pm$ #2}

\newcommand{\picThreePlus}[6]{
	\begin{figure*}[tb]
		\centering
		\begin{subfigure}[c]{0.3\textwidth}
			\centering		
			\includegraphics[width=0.9\textwidth]{#1-0}
			\subcaption{#2}	
				\label{fig:#1-0}	
		\end{subfigure}
		\begin{subfigure}[c]{0.3\textwidth}	
			\centering	
			\includegraphics[width=0.9\textwidth]{#1-1}
			\subcaption{#3}	
				\label{fig:#1-1}	
		\end{subfigure}
		\begin{subfigure}[c]{0.3\textwidth}	
			\centering	
			\includegraphics[width=0.9\textwidth]{#1-2}
			\subcaption{#4}	
				\label{fig:#1-2}	
		\end{subfigure}
		\begin{subfigure}[c]{0.9\textwidth}	
			\centering	
			\includegraphics[width=0.9\textwidth]{#1-3}
			\subcaption{#5}	
				\label{fig:#1-3}	
		\end{subfigure}
	
		\caption{#6}
		\label{fig:#1}
	\end{figure*}
}

\newcommand{\tblAnnotation}{
\begin{table}[tb]
		\caption{
		The first three rows describe the experiment in each row.
		An x marks the usage of a clean seed and --- means not applicable because no support predictions were used.
		The abbreviations for the other columns are defined in \autoref{subsec:metric}.
		The best result per column is marked bold.
	}
	\resizebox{\textwidth}{!}{
		\begin{tabular}{l c c  c c c c c c }
			\toprule
			 & & & \multicolumn{6}{c}{Scores} \\
			  \cmidrule(r){4-9}
			\makecell{Method for\\support predictions} & \makecell{Valid\\Seed} & \makecell{Used\\Output} & \makecell{$\kappa$  $\uparrow$ \\ \mes{\%}}  & \makecell{Acc.  $\uparrow$ \\ \mes{\%}}   & \makecell{F1  $\uparrow$ \\ \mes{\%}} & \makecell{T.Acc.  $\uparrow$ \\ \mes{\%}} & \makecell{Cons.  $\uparrow$ \\ \mes{\%}}  & \makecell{Time $\downarrow$ \\ \mes{min}} \\
			
			\midrule
			
			None & & None & \s{71.35}{2.57} & \s{84.53}{1.23} & \s{80.23}{2.05} & 77.21 & --- & \s{13.95}{2.25} \\
			FOC \cite{foc} & & $p_n$ & \s{48.14}{11.9} & \s{71.78}{7.37} & \s{64.48}{8.25} & 59.24 & 65.15 & \s{9.09}{0.57} \\
			FOC \cite{foc} & & $p_o$ & \s{56.4}{8.67} & \s{73.76}{5.55} & \s{68.69}{6.54} & 62.85 & 69.66 & \s{7.08}{0.44} \\
			S2C2 \cite{s2c2} & x & $p_n$ & \s{79.95}{1.03} & \s{87.92}{0.56} & \s{85.66}{1.14} & 82.01 & \textbf{85.26} & \textbf{\s{5.15}{0.60}} \\
			S2C2 \cite{s2c2} & x & $p_o$ & \s{74.82}{0.17} & \s{84.15}{0.15} & \s{83.57}{0.24} & 76.74 & 84.96 & \s{5.51}{0.32} \\
			S2C2 \cite{s2c2} & x & $p_n$ \& $p_o$ & \textbf{\s{83.17}{2.16} }& \textbf{\s{89.59}{1.29}} &\textbf{ \s{88.08}{1.62}} & \textbf{84.58 }& 84.70 & \s{5.18}{0.63} \\
			
		\end{tabular}
	}
	\label{tbl:annotations}
\end{table}
}

\title{Life is not black and white - Combining Semi-Supervised Learning with fuzzy labels}

\author{ \href{https://orcid.org/0000-0002-6945-5957}{\includegraphics[scale=0.06]{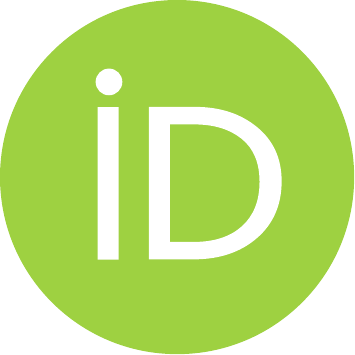}\hspace{1mm}Lars Schmarje}\thanks{Corresponding Author} \\
	Multimedia Information Processing Group\\
	Kiel University\\
	24118 Kiel, Germany \\
	\texttt{las@informatik.uni-kiel.de} \\
	\And
	\href{https://orcid.org/0000-0003-4398-1569}{\includegraphics[scale=0.06]{orcid.pdf}\hspace{1mm}Reinhard Koch} \\
	Multimedia Information Processing Group\\
	Kiel University\\
	24118 Kiel, Germany \\
	\texttt{rk@informatik.uni-kiel.de} \\
}

\renewcommand{\headeright}{A concept paper}
\renewcommand{\undertitle}{A concept paper}
\renewcommand{\shorttitle}{Life is not black and white}

\hypersetup{
pdftitle={Life is not black and white - Combining Semi-Supervised Learning with fuzzy labels},
pdfauthor={Lars Schmarje},
pdfkeywords={Semi-supervised, ambiguous labels},
}

\begin{document}
\maketitle

\begin{abstract}
The required amount of labeled data is one of the biggest issues in deep learning.
Semi-Supervised Learning can potentially solve this issue by using additional unlabeled data.
However, many datasets suffer from variability in the annotations.
The aggregated labels from these annotation are not consistent between different annotators and thus are considered fuzzy.
These fuzzy labels are often not considered by Semi-Supervised Learning.
This leads either to an inferior performance or to higher initial annotation costs in the complete machine learning development cycle.
We envision the incorporation of fuzzy labels into Semi-Supervised Learning and give a proof-of-concept of the potential lower costs and higher consistency in the complete development cycle.
As part of our concept, we discuss current limitations, futures research opportunities and potential broad impacts.
\end{abstract}

\section{Introduction}

Deep Learning was successfully applied to many computer vision problems over the last years.
One of the biggest issues with deep learning is the required amount of labeled data for training.
Thus, many Semi- and Self-Supervised algorithms have been proposed which can decrease the required labeled data by using additional unlabeled data \cite{simclr,simclrv2,fixmatch,s2c2,byol,survey}.
These methods aim for the clear vision that we feed the complete data into the network and additionally provide only few labeled samples per class.
This step can then be used in a general machine learning development cycle  (\mldc, \autoref{fig:idea-0}) to reduce the annotation cost.
We describe the \mldc as a three step cycle (data collection, model training and model evaluation) based on \cite{ml-pitfalls}.

However, this vision is missing two important facts because life is not just black and white.
Firstly, we as humans have to provide such labels.
We will make mistakes and suffer from inconsistency in the form of intra- and interobserver variability \cite{merIssue,noisy-labels-comparison}.
This means that the labels we provide may be wrong or even differ over time or between annotators.
Secondly, in the real-world, we often encounter an ambiguous situation where a true label is either difficult to obtain or not existing.
For example, the cross bread of two different dogs can not be classified as one of its parents.
These two issues have been summarized before as \emph{fuzzy} labels \cite{foc,s2c2} and have been discussed as problematic for many  Semi-Supervised Learning algorithms.
The issue of fuzzy labels is common in many life science \cite{planktonUncertain,tailception,benthic_uncertainty}  or medical  \cite{mammo-variability} datasets.
Even in curated datasets like Imagenet \cite{imagenet} and CIFAR10 \cite{cifar,cifar10h} these issues can exist for example due to correlated label noise with visual similar classes \cite{correlated_label_noise}.
In current research, these issues are often countered by expensive label cleaning \cite{are_we_done} or with noise estimation \cite{divide-mix}.
We believe these perspectives are often too narrow and lead to unnecessary prerequisites of other steps in the \mldc and/or are expensive with regard to the required annotation time.
In this paper, we will give an alternative perspective of incorporating Semi-Supervised Learning into a \mldc which is aware of the issue of fuzzy labels (\autoref{fig:idea-2}).

\subsection{The issue of fuzzy labels}

A \emph{fuzzy} label $l$ is a probability distribution over the classes $C$ for an image $x$.
A label is the aggregation of the annotations $a_1$, ... , $a_n$ for $x$. 
These annotations are one-hot encoded estimates of the label $l$ by human annotators.
Fuzzy labels are aggregated annotations which are not consistent with each other (e.g. $\forall i : a_i \neq \frac{1}{n} \sum_j a_j $).
For example, if you have 10 annotations and 5 annotations are for class A while the other 5 are for class B, the aggregated fuzzy label could be 0.5 for each class.
If all annotations are consistent with each other, we call the label \emph{\certain}.
This definition is similar to soft and hard labels.
However, every hard label could also be represented by a one-hot encoded soft-label but a label is either fuzzy or \certain.
Throughout this paper, we will call also the corresponding image $x$ fuzzy or \certain depending on its label.
A dataset is called \certain if all images are \certain otherwise it is fuzzy.

These fuzzy labels pose two issues we need to overcome to achieve the vision of Semi-Supervised Learning.
The first issue is that we need to incorporate the knowledge about the existence of fuzzy labels into the training.
Many Semi-Supervised Learning (SSL) algorithms assume that only \certain labels exist in their dataset. 
Schmarje et al. \cite{foc} showed that SLL algorithms have a inferior performance if they do not consider the existence of fuzzy labels.
An assumption for the reason for this performance was that fuzzy images are not easily classifiable into the existing classes and thus confuse the algorithm.
The fuzzy labels can be incorporate in the training procedure with for example S2C2 which is an extension to most existing SSL algorithms \cite{s2c2}.
The second issue is that most algorithms aim at good hard classifications as output for the neural network. 
This hard classification makes sense for \certain labels but does not work properly for fuzzy labels as they may have no best hard label.
If we want to integrate the output of an algorithm into a \mldc, we need to consider this.
For example, it could be necessary to resort fuzzy data either into an existing class, a new class or exclude the data.

\subsection{Concept}

Our main concept is illustrated in \autoref{fig:idea-2}.
We will first repeat the issue of most Semi-Supervised Learning (SSL) in the \mldc on uncurated data and than show the difference to our idea.

As stated above, SSL aims at reducing the annotation cost by leveraging unlabeled data which is often expected to be \certain or resulting in inferior performance \cite{foc}.
However, if we have a uncurated dataset, we argue that the requirement of \certain labels is not given. 
It is costly to detect these \certain labels in the given dataset and determine their label.
The costs are so high because we need to use strategies e.g. a consensus process or strict protocols like \cite{ideaProtocol} to counter intra- and interobserver variability.
Moreover, we need a post-processing to decide what we do with difficult, fuzzy images. 
If we have for example an image with a label of 50\% for two classes we have to select a  hard label or ignore this image.
After this costly dataset cleaning, we can apply any SSL algorithm with a relative low cost (\autoref{fig:idea-1}). 
The generated predictions can than be used to annotate more data more easily.
The main issue is that the SSL algorithms expect also cleaned unlabeled data e.g. no fuzzy intermediate images between classes as \cite{foc} hypothesises.
Otherwise, some main assumptions like the separation of classes in a higher feature space might not be valid.

We envision a machine learning cycle that includes the knowledge of fuzzy labels into the SSL algorithm (\autoref{fig:idea-2}). 
Thus, the initial annotation could then be limited to a small portion of the data which we leverage as labeled data.
No assumptions and annotations are then need for the unlabeled data.
This could for example be achieved by additional clustering of the data or an automatic distinction between \certain and fuzzy images.
A consensus step could still be required in the \mldc to use these output predictions as additional input.
But this step could be implemented based on predictions of the network and thus make the annotation and consensus cheaper.
Overall, we expect a lower number of required annotations and thus cost while achieving a higher consistency in the data for our envisioned development cycle in comparison to the previous one.
We give a proof-of-concept in the next section.

\picThreePlus{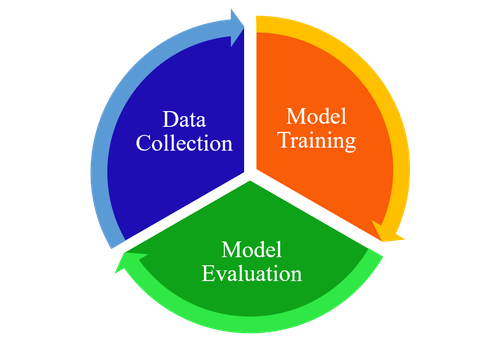}{Machine Learning Development Cycle}{SSL}{SSL + Fuzzy Labels}{Mice Bone Example Images}{
The left image (a) illustrates a simplified machine learning development cycle. The middle (b) and right (c) image how this process can be visualized with Semi-Supervised Learning (SSL) without and with the consideration of fuzzy labels.
The light gray circles represent unlabeled images.
The colored circles represent labeled images or the prediction of these labeled images.
Dark gray circles represent datapoints which are identified as fuzzy.
The number of thunderbolts indicates the number of required annotations by humans in each step. 
In the lower figure (d), we show 9 real-world Mice Bone examples from \cite{s2c2}. 
The images of the classes similar (green) and dissimilar (orange) fiber orientations are easy to consistently annotate.
The middle images (grey) are more difficult to consistently annotate and thus are fuzzy.
}

\section{Proof-of-Concept}

\subsection{Dataset}
\label{subsec:dataset}

We use the Mice Bone dataset proposed in \cite{s2c2} and show examples in \autoref{fig:idea-3}.
This dataset consists of gray-scale images of collagen fibers of mice bone.
The classification problem has three classes: similar, dissimilar and not relevant fiber structures.
We follow \cite{s2c2,foc} and use only \certain images for training and validation and enforce a class balance in this data.
We call this \certain and balanced training data \emph{seed} and use the rest of the data as unlabeled data.

We use two different seeds for the Mice Bone data.
The first seed is based on the segmentation masks from \cite{schmarje2019}. 
Due to the dimension reduction from a segmentation to a classification label and the uncertain segmentation, the labels show high variability and only one annotation is available. 
We treat all labels as \certain and this leads to an inconsistent seed.
The second seed is based on three independent annotations of each image. 
These annotations allow us to estimate the fuzziness of each image and thus lead to a valid seed.

\subsection{Metrics}
\label{subsec:metric}

We aim at faster and more consistent annotations throughout the complete \mldc.
Every presented experiment is executed on the same raw data with one of above-described seeds. 
An experiment consists of three independent annotations of the same person over time for the complete dataset.
For ease of referencing, we view these independent annotations over time as three (intra-)annotators.
The annotations were either done without any additional information or with different outputs of an SSL algorithm given as \emph{support (predictions)}.
These support predictions could be accepted or manually corrected.
This support can also be an overclustering of the data. 
An overclustering is a clustering with more cluster than used classes \cite{foc,s2c2}.
We give in total 6 different evaluation metrics which are either average over the three annotators or the cross-combinations between them.

Cohen's \emph{kappa} coefficient ($\kappa$) is a statistical metric that is often used to measure the intra- and inter-observer variability \cite{kappa}.
The coefficient measures the agreement between two annotators for a classification task.
\emph{Accuracy} (Acc.) measures the true positives in relation to all annotations between two annotators.
\emph{F1-Score} aggregate the precision and recall between the annotation of two annotators. 
\emph{Total Accuracy} (T.Acc.) counts the consistent annotations between all annotators and divides them by the total number.
This metric is like Acc. but not for a pair of annotators but for all three annotators in parallel.
\emph{Consistency} (Cons.) divides how many predictions were considered consistent by the total number of images.
An image is consistent with the given support predictions, if no manual correction was applied.
\emph{Time} is the time it took the annotator to label the complete dataset.

\subsection{Results}
\label{subsec:results}

We discuss six different experimental setup with support predictions as proof-of-concept results in \autoref{tbl:annotations}.
The first experiments used no support predictions. 
These results of this experiment were used to calculate the above-mentioned clean seed.
The second and third experiments used support predictions from the inconsistent seed with the method FOC \cite{foc}. 
Either a classification head $p_n$ or an overclustering head $p_o$ was used as output resulting in a classification or overclustering of the data respectively.
The last three experiments used a support calculated on the clean seed with the method S2C2 \cite{s2c2}.
As output either classification head $p_n$, a overclustering head $p_o$  or both heads ( $p_n$ \& $p_o$)  based on the fuzziness estimation were used.

\tblAnnotation

We want to highlight three important aspects.
Firstly, the usage of support reduces the required annotation time while not necessarily improving its consistency.
Secondly, the consistency with support is worse with an inconsistent seed and the method FOC in comparison to the others.
Due to the similar architecture of FOC and S2C2, we attribute this worse performance rather to the inconsistent seed than to the method.
Thirdly, S2C2 (with both outputs) more than halves the required annotation time while increases almost all other metrics.

\section{Discussion}
\label{sec:discuss}

\subsection{Future Research}
Our proof-of-concept experiments show that using the output SSL algorithms that are aware of fuzzy labels can improve the consistency of acquired labels and reduce their annotation time.
However, this benefit is not always achieved when using network predictions which enforces the importance of careful seed selection and/or SSL algorithm.
As a first proof-of-concept, these results lack additional experiments such as different cross-combinations and comparisons to noise estimation  and other algorithms.
While these issues have to be addressed in future research, the results illustrate the potential benefits of our concept for a \mldc that considers fuzzy labels.
Future research could also incorporate  uncertainty estimates from humans and neural network predictions into the annotation process.
A promising research direction would also be large user studies and the simulation of such studies for detailed ablations.

\subsection{Broader Impact}

Semi-Supervised Learnig (SSL) aims at decreasing the required amount of labeled data to a few samples per class.
We envision a \mldc which allows us to apply SSL to most classification problems with relatively low cost.
Especially, in domains like medical imaging, the required labeled data is a severe limiting factor. 
If we could resolve this issue with SSL which considers fuzzy labels, we would open a vast variety of new research opportunities at a large scale which is currently not feasible.

If we incorporate model predictions into our annotation process, we need to be aware that we might suffer from a confirmation bias.
In the worst cases scenario, we could create a self-fulfilling prophecy which would lead to a degeneration of the complete \mldc.
However, by carefully monitoring our processes as part of the development cycle we can detect these issues early.
Additionally, we assume that cautiously leveraging this bias can lead to more high-quality data which can improve future algorithms and research.

Moreover, we advocate a change of evaluation perspective. 
While it is important to evaluate on highly curated datasets for algorithm development, we also have to look at the complete \mldc.
Otherwise we can not detect the issues described in this paper that fuzzy labels are a limiting factor for applying SSL to different real-world data.
We already see great benefits when applying deep learning to other research fields but decreasing such limiting factors further can potentially increase the impact of deep learning even more.

\section{Conclusion}

Semi-Supervised Learning (SSL) has great potential by solving one big issue of deep learning: The required amount of labeled data and its cost.
However, many SSL algorithms do not take fuzzy labels into account and thus require expensive prepossessing steps of the data.
By including the knowledge of fuzzy labels into our machine learning development cycle (\mldc), we envision that we can decrease the overall annotation time and thus its costs while achieving a higher consistency.
We give a proof-of-concept of this concept on a Mice Bone dataset with fuzzy labels.
We highlight that a lot of future research opportunities exist to validate and improve the presented ideas.
We advocate to broaden our view from the algorithm development to the the complete \mldc to detect and overcome issues like fuzzy labels.
This could spark a variety of before-infeasible research opportunities and thus lead to new breakthroughs in science in general.

\bibliographystyle{IEEEtranN}
\bibliography{lib}

\end{document}